\documentclass[runningheads]{llncs}
\usepackage{amsmath}
\usepackage{amssymb}
\usepackage{svg}
\usepackage{graphicx}
\usepackage{hyperref}
\usepackage{url}

\begin{document}
\title{Large Language Models for Difficulty Estimation of Foreign Language Content with Application to Language Learning}


\author{Michalis Vlachos\inst{1}\and
Mircea Lungu \inst{2} \and \\
Yash Raj Shrestha\inst{1} \and
Johannes Rudolf David \inst{1}
}

\institute{Department of Information Systems, HEC, University of Lausanne \and
IT University of Copenhagen
}

\titlerunning{LLMs for Difficulty Estimation of Foreign Language Content}

\authorrunning{Vlachos, Lungu, Shrestha, David}


\maketitle          
\begin{abstract}
We use large language models to aid learners enhance proficiency in a foreign language. This is accomplished by identifying content on topics that the user is interested in,  and that closely align with the learner's proficiency level in that foreign language.
Our work centers on French content, but our approach is readily transferable to other languages. Our solution offers several distinctive characteristics that differentiate it from existing language-learning solutions, such as, a) the discovery of content across topics that the learner cares about, thus increasing motivation, b) a more precise estimation of the linguistic difficulty of the content than traditional readability measures, and c) the availability of both textual and video-based content. The linguistic complexity of video content is derived from the video captions. It is our aspiration that such technology will enable learners to remain engaged in the language-learning process by continuously adapting the topics and the difficulty of the content to align with the learners' evolving interests and learning objectives.

\medskip
A video showcasing our solution can be found at:\\ \url{https://youtu.be/O6krGN-LTGI}

\keywords{digital education \and competency-based learning \and extensive reading 
 \and machine learning 
\and large language models }
\end{abstract}

\section{Introduction}

The value of gaining communicative competence is widely acknowledged, especially in today's interconnected world. Whether it is for expanding one's social circle, finding a job, as a hobby, or even as a protective mechanism against cognitive decline \cite{10.3389/fnhum.2018.00305}, learning a foreign language is a rewarding but also challenging task. 
A major hurdle pertains to the time required to build up the new vocabulary and to use the novel linguistic patterns in a fluid manner. 
Another challenge arises from the fact that learning a new language via the traditional way of using textbooks is a non-personalized and often unengaging process. This is because textbook content is static and follows a linear structure, and cannot adapt to the learner's changing interests, requirements, and level of proficiency in the foreign language.

Contemporary online foreign-language educational tools, such as \href{https://www.duolingo.com}{Duolingo}, \href{https://www.gymglish.com/fr/frantastique}{Frantastique}, \href{https://readlang.com}{ReadLang}, and others, integrate elements of gamification into the learning process to enhance their appeal and to ultimately decrease learner attrition from their learning platforms \cite{tuncay2020app}.

However, existing tools still lack important personalization aspects because they either 
    work with predefined generic materials (Duolingo, Frantastique), or 
    expect the user to manually search for their own textual content on the internet (ReadLang).
We provide a solution that addresses several of the aforementioned shortcomings. Our approach involves the identification of content that is both comprehensible and engaging to users, with the ultimate goal of enhancing their learning journey. Specifically, within this solution, our objective is to pinpoint video and textual material that meets two key criteria: a) it is likely to pique the reader's interest, and b) its linguistic complexity aligns with the learner's current skill level. By seamlessly integrating these two facets, we aim to develop systems that prove highly effective in facilitating foreign language acquisition. Furthermore, we propose the utilization of a recommendation engine, which can suggest matched texts to users based on a combination of difficulty and personal interest.

This endeavor builds upon the ongoing discourse surrounding the integration of AI agents, such as machine learning systems, chatbots, and conversational agents, into the realm of foreign language learning \cite{jeon2021exploring,lin2021learning,underwood2017exploring}. These agents play a pivotal role in supporting language learners by delivering personalized content to individuals and offering ample opportunities for independent language practice \cite{dizon2020evaluating,moussalli2020intelligent}. 

Our solution encompasses machine learning topics such as difficulty estimation, topic prediction, and content recommendation. In this paper, we specifically focus on the methodology of extracting content from the internet and for difficulty prediction of that content. 

In contrast to existing systems that create and utilize their own content, our approach leverages the vast resources of the Internet to discover video and text-based materials that are highly engaging to the learner. To accomplish this, we employ machine learning techniques to gauge the linguistic ``complexity" of each content unit, essentially assessing the proficiency level required in the foreign language to comprehend its vocabulary. Over time, our system is anticipated to enhance its ability to gauge both the user's preferences, thereby recommending topics aligned with the user's interests, and the user's proficiency in the foreign language. Consequently, it will only suggest content that corresponds to the learner's skill level or slightly exceeds it.

From a pedagogical perspective, our solution builds upon the already established theory of \textit{extensive reading}. A large body of research both by Day and Bamford \cite{daybamford2002}, as well as by Krashen \cite{krashen2003explorations}, has shown that extensive reading constitutes a crucial means of reinforcing one's language skills, not only improving reading and vocabulary skills, but also yielding more comprehensive improvements across all areas of language competence \cite{daybamford2002,krashen2004power}. Alan Maley, a distinguished English scholar, has even advocated extensive reading as the ``single most important way to improve language proficiency" \cite{maley2005}. 

From a technological perspective, our work makes the following \textbf{contributions}:

1. We provide a solution using modern machine learning techniques to estimate the \textbf{difficulty} of digital content. Our method is based on modern text embeddings and large language models to estimate the text difficulty. Furthermore, our solution identifies content on topics that the learner is likely to be interested in.

2. Extant research primarily addressed books or articles, but our solution is more comprehensive in that it retrieves also available \textbf{video content} from YouTube, and uses automatic captioning to infer its linguistic difficulty and topic. In this manner, our approach benefits from the broad availability of engaging content in video format, and thus, it is does not require extensive human resources to create manually new educational content.




\section{Pedagogical Underpinnings}

Our learning approach is rooted on several well-established pedagogical principles. The first one being extensive readings (also known as Free Voluntary Reading \cite{krashen2004power}) which encourages students to choose their own reading materials and read for pleasure. Essentially, it asserts that the pleasure derived from reading keeps learners engaged with the content and reduces attrition from the learning program.
A list of key principles for successful extensive reading has been proposed by \cite{daybamford2002} and then extended by \cite{prowse2002top}  which we reiterate in Figure 
\ref{fig:extensivereading}.

\medskip \noindent

\begin{figure}
    \centering
    \noindent\fbox{%
    \parbox{\linewidth}{%
\begingroup
\fontsize{8pt}{12pt}\selectfont
   
1. Students read a lot and read often.\\
2. There is a wide variety of text types and topics to choose from.\\
3. The texts are not just interesting: they are engaging/compelling. \\
4. Students choose what to read.\\
5. Reading purposes focus on: pleasure, information and general understanding.\\
6. Reading is its own reward.\\
7. There are no tests, no exercises, no questions and no dictionaries.\\
8. Materials are within the language competence of the students.\\
9. Reading is individual, and silent.\\
10. Speed is faster, not deliberate and slow.\\
11. The teacher explains the goals and procedures clearly, then monitors and guides the students.\\
12. The teacher is a role model…a reader, who participates along with the students.\\
 
\endgroup
}}
    
    \caption{Principles of extensive reading}
    \label{fig:extensivereading}
\end{figure}

The reading workshop model \cite{atwell2015middle} is a related approach that involves learners reading a lot on topics they find interesting. This approach allows students to choose books based on their own interests and then participate in book talks, book clubs, and other reading-based activities to encourage discussion and analysis of the texts they have read. Project-based learning is a different approach where learners are encouraged to select a topic they are interested in and read extensively on it as part of the project \cite{boss2022reinventing}. This approach emphasizes the development of skills such as critical thinking, problem-solving, and collaboration while allowing learners to pursue their own interests and passions through reading.

Competency-based learning \cite{burnette2016renewal,book2014all} is an educational approach that focuses on the mastery of specific skills or competencies, rather than just completing a certain number of courses or hours of study. Students are empowered daily to make important decisions about their learning experiences, how they will create and apply knowledge, and how they will demonstrate their learning. Students receive timely, differentiated support based on their individual learning needs, and they learn actively using different pathways and varied pacing.

\medskip
As we elaborate on our solution, it will become clear to the reader that our approach directly addresses the first ten principles of extensive reading and is drawing upon several tenets of the aforementioned pedagogical approaches: we discover content that the learner cares about, reading/studying is done as a pleasure, the content is differentiated per learner and is selected to be within the learner's language competence.

\section{Our approach and technological underpinnings}
We describe our methodology and the architecture of our solution.

\begin{figure}[!h]
\centering
    \includegraphics[width=0.8\linewidth]{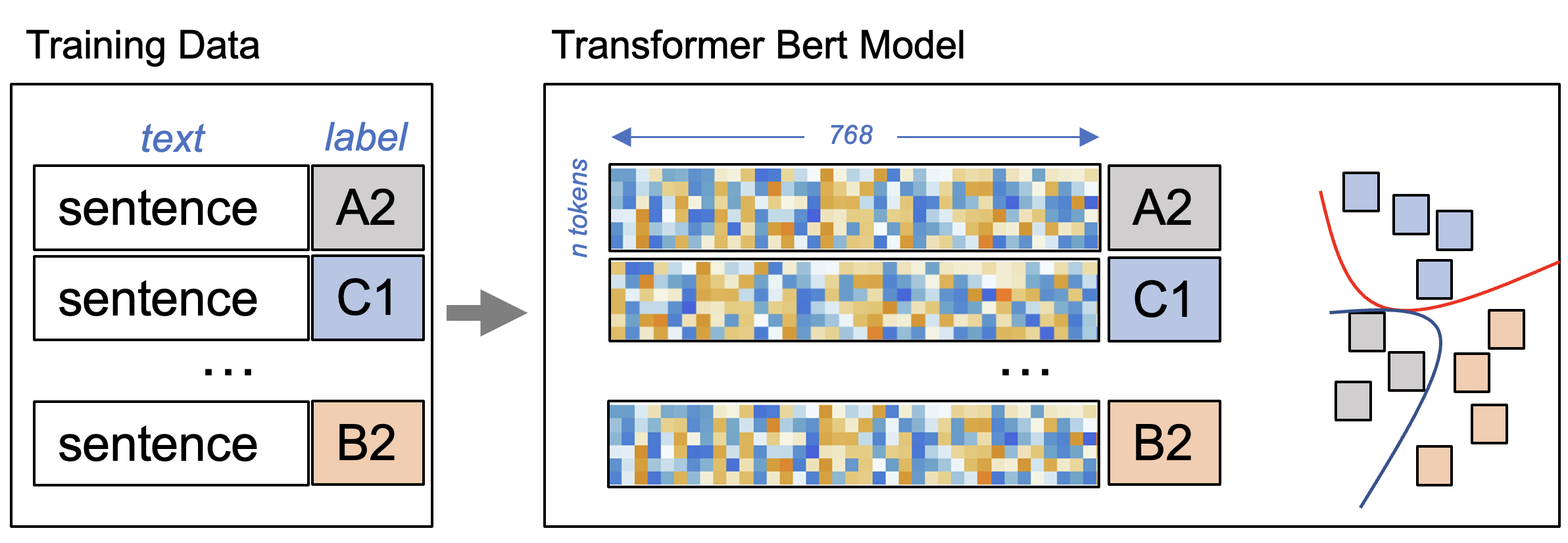}
    \caption{Posing difficulty estimation as a classification problem.}
    \label{fig:bert}
\end{figure}

\medskip
\noindent
\textbf{Estimating Difficulty/Level of text:}
We model the estimation of difficulty as a classification problem. So, if \(\mathcal{D}\) is the set of documents to be classified, and let \(Y\) be the random variable representing the linguistic difficulty class. For example, if we label difficulty based on the CEFR levels, \(Y\) can take values from the set \(\{A1, A2, B1, B2, C1, C2\}\).  A classifier is a function 

$$f: \mathcal{D} \rightarrow Y$$

\noindent
that maps a document \(d \in \mathcal{D}\) to a linguistic difficulty class \(y \in Y\).
The classifier $f$ is built using pairs of text and a label, where the label corresponds to the linguistic difficulty of the content. To create this model that predicts the difficulty of foreign language content, we use modern transformer neural networks and large language models for which there exist several instances, e.g., BERT \cite{Bert2019}, GPT \cite{GPT}, GPT-3 \cite{GPT3}, LLaMa \cite{llama}, Palm \cite{palm}, and other. These models vary primarily in terms of the volume of data utilized and the training methodology employed, resulting in a models of diverse size and language capabilities.
As an example, a language model such as BERT will convert the tokens of a text sentence into a series of 768-dimensional vectors (also known as \textit{embeddings}) capturing the ``meaning" of each token (see Figure \ref{fig:bert}). Use of the different models will result into vectors of different lengths. For example the ``ada-002" embeddings of OpenAI produce vectors of length 1536.  They also consider a context of more than 8000 tokens, thus they can encode more accurately each word based on its surroundings.


\begin{figure}[!h]
\centering
    \includegraphics[width=0.45\linewidth]{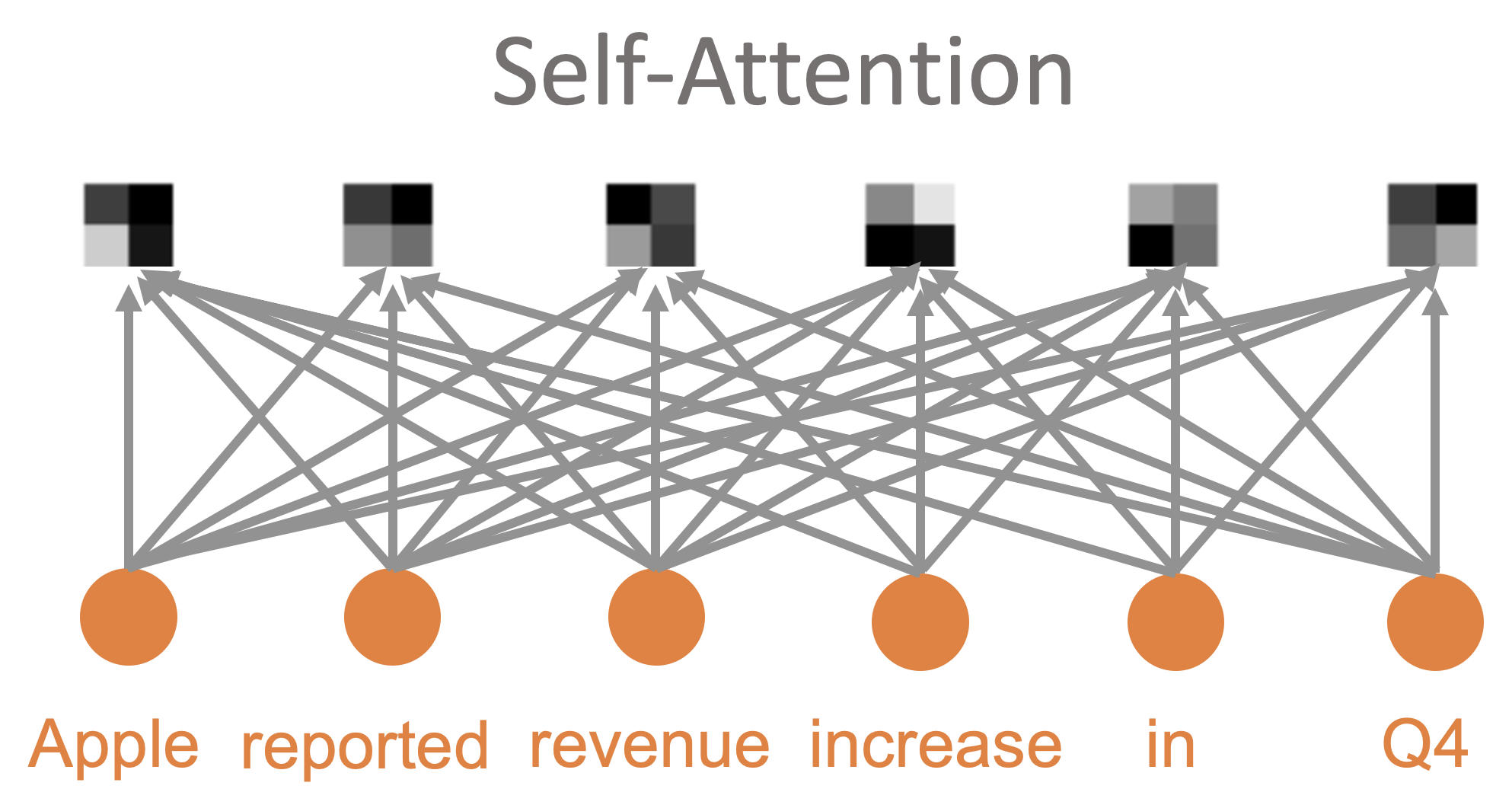}
    \caption{Illustration of self-attention used by large language models}
    \label{fig:attention}
\end{figure}

The capability of large language models to capture the context of each word by considering the neighborhoods of words and sub words gives them tremendous linguistic power. They accomplish this by capitalizing on a self-attention mechanism, as illustrated in Figure \ref{fig:attention}. Earlier embedding techniques such as Glove \cite{pennington2014glove} and Word2Vec \cite{mikolov2013word2vec} did not consider the context and thus create embeddings of inferior quality. As an example of a contextual embedding, if we have three sentences, two of them talking about Apple the company and one referring to the fruit, the texts that refer to the company will be mapped closer to each other, thanks to the disambiguation provided by the context. 

Note, that the original embeddings are only used as seeds for the final model. The neural network model that is created will be post-trained for several epochs using the new training labels provided. This process, also called ``transfer-learning", leads to a more efficient training and creation of the new neural network because we do not need to start from scratch, but we tune further the model by adjusting to the goals of the target task. As a result, this process not only reduces the training time but also requires fewer resources (energy and compute) to train the machine learning model. 

One shortcoming of large language models, as their name suggests, is their large size. For example, models based on the BERT transformer model have sizes in the order of 500MB, while a modern model such as on LLaMa may require more than 100GB of space. One can further compress the model created using quantization and pruning techniques, a task also known as ``knowledge-distillation" \cite{gou2021knowledge}. For example, there exist compressed versions of popular models, such as the TinyBERT \cite{jiao-etal-2020-tinybert} or DistilBert \cite{deepsparse} for the BERT model. There is an increasing number of repositories that offer distilled version of already created models, one example being SparseZoo\footnote{\url{https://sparsezoo.neuralmagic.com}}. Such compressed models can be up to 10 times smaller than the original models, and in practice may also lead to improvements in prediction accuracy, because of the inherent regularization offered via the compression process.

\medskip
\noindent
\textbf{GPT-3+ Models:} During the last years, even more advanced large language models have been introduced in the Natural Language Processing (NLP) literature. Very prominent is the Generative Pre-trained Transformer 3 and 4 by OpenAI, or GPT-3 and GPT-4 for short \cite{NEURIPS2020_1457c0d6}, which have been trained on massive datasets of multilingual text to learn patterns and relationships in text. These advances have led to the development of the popular ChatGPT, a conversational AI capable of engaging with users and producing a wide spectrum of content, spanning from poetry to programming code. GPT-3 and GPT-4, have been pre-trained for next-token prediction and are post-trained using reinforcement learning from human feedback. GPT-3 and GPT-4 models achieve state-of-the art performance across a broad array of tasks including text summarization, translation and question-answering \cite{bubeck_sparks_2023}. Such models are particularly good at generating human-like text, but they can also be used as a basis for building advanced classification models, particularly with post-training (often referred to as fine-tuning) on data labelled for the task at hand. In our experiments, we also use the GPT-3 model (and not the newer GPT-4) because the OpenAI platform allows for fine-tuning, at the current time point, only on the GPT-3 model. In the experiments, we see that the GPT-3-based model builds the most accurate difficulty predictors given the labelled data provided.


\medskip \noindent
 \textbf{Transfer Learning:} Fine-tuning of the large language models is an instance of transfer learning \cite{transferLearning1,transferLearning2,transferLearning3}. In transfer learning, the knowledge gained from one task (source task) is utilized to improve the performance of a related but different task (target task). In the context of text analysis, transfer learning involves leveraging a pre-trained model's learned features and representations from a source text-related task to enhance the performance of a target task. In our case, the task is a classification task which is predicting the difficulty level of a given text. 
 
 If we assume that the pre-trained model has some parameters $\theta$, with fine-tuning we only fully re-train the weights for the last few layers of the (neural network) model which hold the labels. The existing model is post-trained using the existing pre-trained parameters and adapting the loss function so that the model adapts to predict more accurately the new labels and labelled examples.

So, if the target task is a text difficulty prediction, and $x$ is a given text we fine-tune the pre-trained model's parameters using the target task data. The model's parameters $\theta$ are updated based on the loss computed using the predicted difficulty labels and the ground truth labels from the target task data:
   
   $$\mathcal{L}_{\text{target}}(\theta) = \sum_{(x, y) \in \mathcal{D}_{\text{target}}} \text{loss}(\text{pred}(x; \theta), y)$$

\noindent where
$\text{pred}(x; \theta)$ represents the predicted difficulty label for input text $x$ using the model parameters $\theta$.
Typically, one can also balance the loss objectives of the pre-trained large language model (LLM) and the target task with a parameter $\lambda$, eg: 

$$\mathcal{L}_{\text{transfer}}(\theta) = \mathcal{L}_{\text{LLM}}(\theta) + \lambda \cdot \mathcal{L}_{\text{target}}(\theta)$$

The advantage of transfer learning is that it is much faster to train such a model, since the model parameters need not be learned from scratch, but we start with a ``good-enough", generic model, and we make it more focused to a particular task with the provision of specific examples. In such a way, the rich knowledge obtained from large bodies of training data is repurposed with task-specific labelled data.

\medskip
\noindent
\textbf{Topic detection:} Detecting the topics (e.g., politics, sports, movies, etc)  of a digital content (article, video) is necessary for our solution. This ensures that we can align with the users' interests when delivering the content to them. Note, that much of the digital content either is readily available as text, or can be converted to text: e.g., for a video one can extract the captions using speech-to-text technology, for music one can use the lyrics, etc. Topic detection can also be modelled as a classification problem. That is, one can provide several examples of text and their labels with topics that the text refers to. Here, we do not need to create our own model nor do we collect our own data as we did for the difficulty estimation, but we use already pre-trained models based on zero-shot text classification \cite{yin2019benchmarking}. Such solutions can detect a broad spectrum of \href{https://huggingface.co/mjwong/multilingual-e5-large-xnli-anli}{topics}, and have been shown to be very effective, reaching accuracies that exceed $90\%$ in a variety of benchmark datasets. These very effective topic classifiers are trained on multi-lingual datasets and therefore can be directly used for our application. In the case that a topic detection model is trained only on English text, then it is desirable first to translate the content into English before providing it to the topic classifier.

Note, that in some cases, the digital content itself already offers a categorization into topics. For example, many YouTube videos come pre-tagged with topics which can offer a very fine-grained topic classification. If such pre-existing topics are present in the content, they are merged with the topics identified by the topic classifier.

%

\begin{figure*}
    \centering
    \includegraphics[width=1\linewidth]{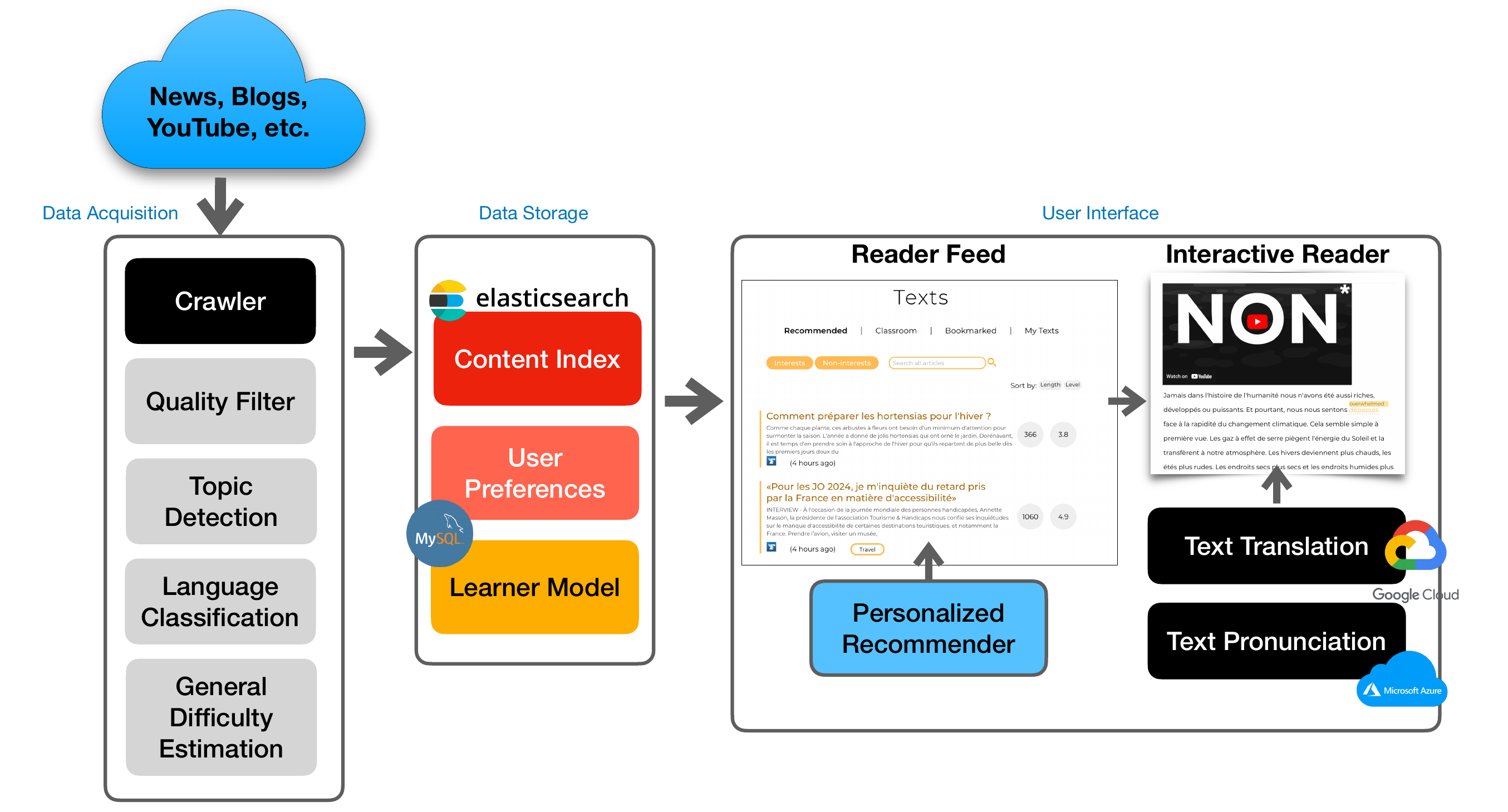}
    \caption{Architecture of our method. Content already available in the internet is acquired and classified into topic, language, and difficulty.}
    \label{fig:architecture}
\end{figure*}

\subsection{Architecture}
The general architecture of our solution is shown in Figure \ref{fig:architecture}, consisting of data acquisition and cleaning components, data annotation components, data storage components, and user interface components. First, 
a {\bf crawler} component monitors a series of RSS feeds (including the RSS of relevant YouTube channels) and every time a new item is discovered it passes it to a {\bf quality filter}. The filter  removes items that are behind a paywall, articles that have too little text, etc.  Several other types of meta-information about the article are extracted by the crawler to be used later in the UI: authors, word count, and a short article description (usually the RSS feed item description is used for this). 

The {\bf topic detection} component extracts one or more topics from a text, both general and user defined. At the same time, the {\bf general difficulty estimation} applies a series of difficulty estimation algorithms on the text, including the difficulty estimation as described in this work and also traditional readability measures \cite{franccois2014analysis}.

The article together with the metadata, topics, and difficulty metrics are then added to the {\bf text index} module which is implemented on top of an Elasticsearch cluster. From there, the {\bf personalized recommender} creates a new mix of fresh recommendations every time the user logs into the web application. It takes into account the {\bf user defined interests} and also the evolving {\bf learner model} which tracks the current knowledge level of a particular user. 

The {\bf interactive reader} provides translation and pronunciation with the help of third-party {\bf machine translation} services for words that the user needs and sends all these interactions together with explicit user feedback on text difficulty back to the {\bf learner model} that updates the estimated knowledge level of the user.

\begin{figure*}
    \centering
    \includegraphics[width=1\textwidth]{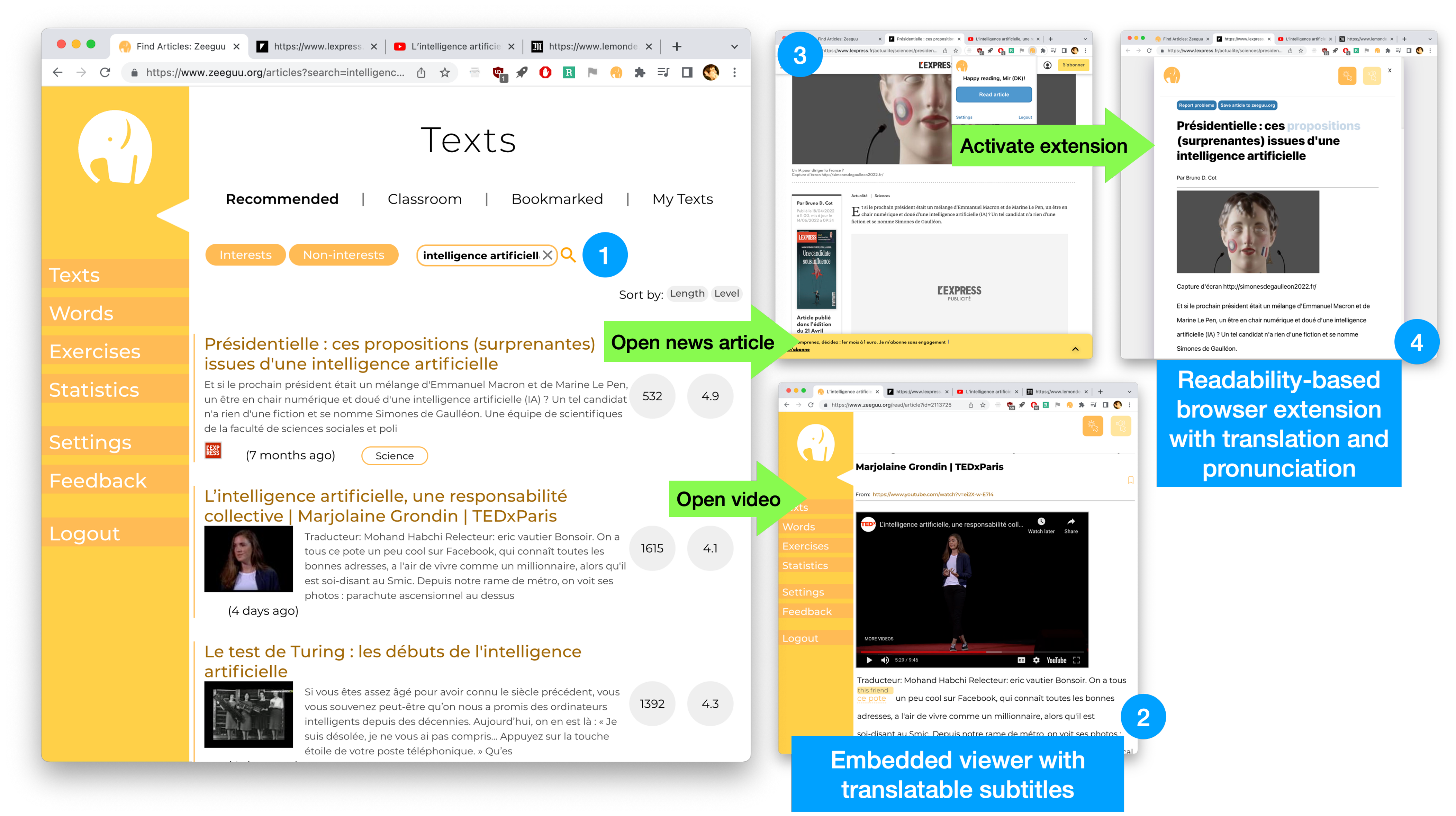}
    \caption{The search results (1) summarize both news and video results. Videos can be watched in an embedded page (2) with translatable subtitles. When an article is being open, the reader is sent to the original source (3) where a Readability-like browser extension is available to clean up the page and support translation and pronunciation in the text (4)}
    \label{fig:UI}
\end{figure*}

\subsection{User interface}

\newcommand{\marker}[1]{\textcircled{\raisebox{-0.9pt}{#1}}}

We integrated the difficulty estimation algorithm presented in the previous sections on Zeeguu, an open source research project \cite{Lungu18}, which provides an easy to deploy platform for offering content to multiple users. 
Figure \ref{fig:UI} presents the main components of the UI. The landing page of the application presents the recommended content for the learner, allowing the customization of the user interests. Users can declare both ``interests`` and ``non-interests``. In Figure \ref{fig:UI}, the user has queried (marked with \marker{1}) the  phrase ``intelligence artificielle'' (i.e.,  Artificial Intelligence in French) and the system has returned a series of videos and articles. Upon opening an article, the user is sent to the original page of this article (\marker{3}). If the learner also has the related reader extension installed on their browser, they can activate it (\marker{4}) to remove the ads and the navigation from the page and focus only on the text which becomes interactive: words and phrases can be translated and pronounced.  
If, on the other hand, the user chooses to watch a video, the video is opened inside the web app (\marker{2}) where the subtitles, are presented with interactive translation and pronunciation capabilities.

\section{Experiments}
The goal of the following experiments is to demonstrate that the difficulty estimation based on the fine-tuned LLMs can significantly improve the difficulty estimation offered by traditional readability metrics. We also show that the newer LLMs significantly outperform the older transformer-based language models. We evaluated our techniques on three \textbf{datasets}:

\begin{enumerate}
\item Littérature de jeunesse libre (\texttt{LjL}) which we obtained from \cite{opencorporaFrench2022}. Each content item here contains several sentences and a label (labels: level1, level2, level3, level4).

\item A collection of sentences collected from the Internet (\texttt{sentencesInternet}). Each of these sentences was then annotated by at least three annotators in difficulty levels. Only sentences in which all participants agreed on the difficulty annotation were retained (labels: A1,A2,B1,B2,C1,C2). Here, the labels correspond to the levels designed by the Common European Framework of Reference for Languages (CEFR) \footnote{\url{https://stay.fl-france.com/french-levels/}}.

\item A collection of sentences from literature books (\texttt{sentencesBooks}). Each book was annotated with a difficulty level by a Professor of French. All sentences in that book were then given that label. This process involved an OCR pipeline which could lead to faulty detection of characters, so only the sentences without any errors were retained. (labels: A1,A2,B1,B2,C1,C2).
\end{enumerate} 

\begin{table*}[!h]
\caption{Characteristics of the datasets in our experiments}
    \label{tab:datasets}
    \centering
    \begin{tabular}{|l||r|r|r||c|} 
        \hline
         \textbf{Dataset}                            &  \textbf{Total labels} &  \textbf{Total words} & \textbf{Total characters} & \textbf{Labels} \\ \hline
         \texttt{LjL} \cite{opencorporaFrench2022}   &   2,060    & 334,026 &  1,532,442   & level1,\ldots, level4 \\ 
         \texttt{sentencesInternet} &  4,800  & 85,941 & 421,045 & A1, \ldots, C2 \\
         \texttt{sentencesBooks}    &   2,400  & 56,557 &  281,463     & A1, \ldots, C2 \\
         \hline
    \end{tabular}
    
\end{table*}

The characteristics of these datasets are provided in Table \ref{tab:datasets}.
To train and evaluate our model, we used an 80/20 train-test split, therefore the results that we present are for examples which the model saw for the first time.

\begin{figure*}[!ht]
\centering   
\includegraphics[width=\linewidth]{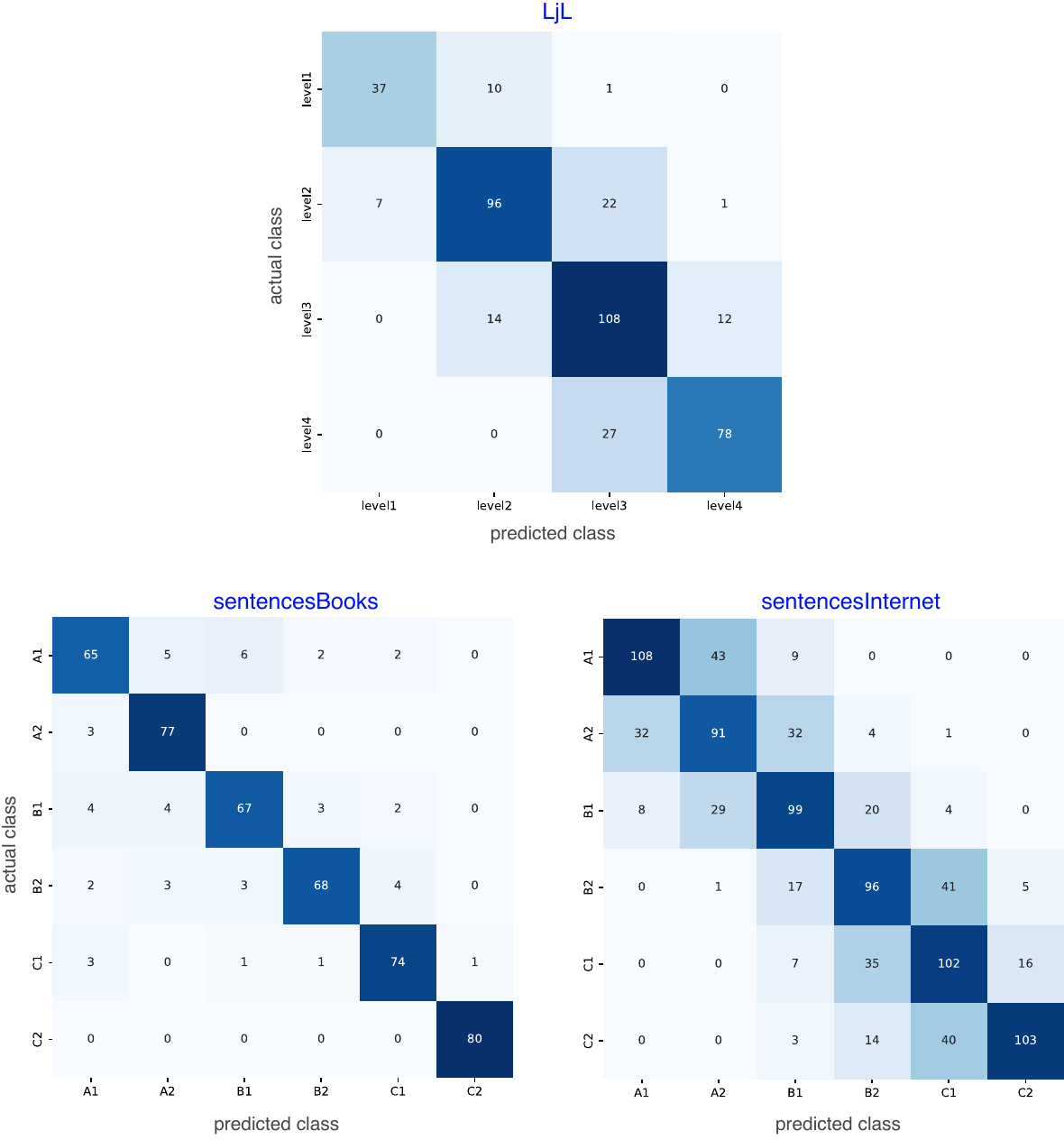}

\caption{The confusion matrix for the difficulty estimation technique based on the GPT-3-based fine-tuned model (DaVinci) for the three datasets in our experiments. It is evident that most errors occur within neighboring class labels, implying that the predictive model seldom makes severe miscalculations.}
\label{fig:CM}
\end{figure*}

\begin{figure*}[!h]
    \centering
\includegraphics[width=\linewidth]{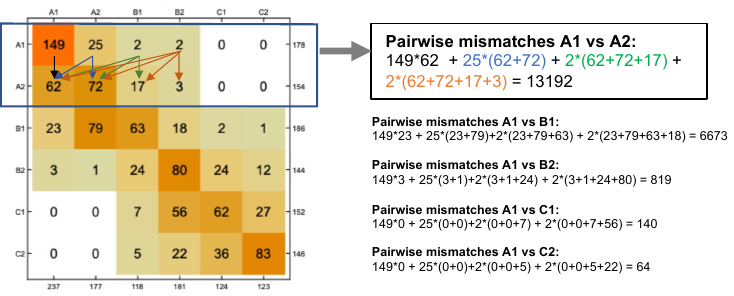}
    \caption{Evaluating the pairwise mismatches for a given confusion matrix of a classifier.}
    \label{fig:pairwiseMismatch}
\end{figure*}

\subsection{Accuracy}
First, as a simple benchmarking, we compare the accuracy of our difficulty estimation approach to traditional readability-based metrics, such as the GFI (Gunning Fog Index), FKGL (Flesch Kincaid grade level), ARI (Automated Readability Index). These techniques are inherently regression techniques and output a floating point value of the text difficulty. As a result,  we cannot make a direct comparison, because our difficulty estimator predicts discrete labels. 

To make a meaningful quality comparison, we devise the following experiment which draws on {\bf pairwise-comparisons}. 
For each text, we predict its difficulty and compare it to all the other objects with different pre-annotated difficulty labels and record if this comparison was correct. For example, if a text with label A1 received an ARI score of 15 and a different sentence with (pre-annotated) label B2 received an ARI score of 14.6 (i.e., indicated as easier), then this comparison is considered incorrect and recorded accordingly. We call this incorrect comparison a \textbf{mismatch}. With our approach, we can also do the same by comparing the labels. If the confusion matrix of a classifier is already computed, then the pairwise mismatches can be evaluated as shown in Figure \ref{fig:pairwiseMismatch}.

The results of this analysis are shown in Table \ref{tab:pairwiseComparisons}. Higher numbers indicate more mismatches. Therefore, we see that compared to older or traditional methods of difficulty estimation, the approaches based on fine-tuning modern transformer neural networks offer a significantly lower error rate. For the GPT-3-based models, there exist several variants. Davinci is the most powerful one, while Curie tries to balance accuracy and speed. Both GPT-3 models have been fine-tuned for the task at hand using the examples with the labelled difficulties of the text. The GPT-3-based model offers a better estimation of difficulty compared to the BERT-based neural network model. Note, that for the BERT model we use the French-based CamemBERT model \cite{martin-etal-2020-camembert}. CamemBERT uses the exact same architecture as BERT, but
the key distinction between them lies in their training data and language focus. While BERT was initially trained on a diverse range of texts in multiple languages, CamemBERT is specifically fine-tuned for the French language. This fine-tuning process involves training the model on a massive amount of French text data, which enables it to capture the nuances, idioms, and syntactic structures unique to the French language.
Because CamemBERT is tailored to French, it excels in various language-related tasks such as text classification, sentiment analysis, and named entity recognition within the context of French text.

\begin{table*}[!h]   
\caption{Number of pairwise mismatches for the neural network difficulty estimation techniques compared to predominant readability metrics. Numbers in bold indicate fewer mismatches, and thus better estimation of difficulty.}
    \label{tab:pairwiseComparisons}
\centering

    \begin{tabular}{|l||c| c |c | l} \hline
 & \texttt{LjL} & \texttt{sentencesBooks} & \texttt{sentencesInternet} &  \\ \hline
\textbf{GFI} & 17,338 & 30,836 & 83,770 &  \\
\textbf{ARI} & 17,749 & 32,993 & 92,140 &  \\
\textbf{FKGL}& 19,634 & 31,475 & 85,306 &  \\ \hline
\textbf{CamemBERT embeddings \cite{martin-etal-2020-camembert}} & 11,354 & 9,875 & 67,288 &  \\
\textbf{GPT-3 Curie fine-tuned} & 11,635 & 7,055 & 60,165 &  \\
\textbf{GPT-3 DaVinci fine-tuned} & \textbf{10,740} & \textbf{6,817} & \textbf{57,258} & \\ \hline
\end{tabular}
    
\end{table*}

We also provide a comparison of the overall prediction accuracy for the transformer based models in Table \ref{tab:accuracy}. The test set contains the same class representation as in the training set. For example, for the \texttt{sentencesInternet} and the \texttt{sentencesBooks} all classes are equi-represented in the train and test set. For the \texttt{LjL} dataset class ``level3" is overrepresented. This is necessary to understand and to show, because we should always compare our classifier with a baseline accuracy, which is always predicting the most represented class. Therefore, for the \texttt{sentencesInternet} and the \texttt{sentencesBooks} the baseline accuracy is $1/6$ or $16.6\%$, whereas for the \texttt{LjL} we have a baseline accuracy of $32.4\%$.

The confusion matrix of the best performing GPT-3 DaVinci model across all three datasets is shown in Figure \ref{fig:CM}. The confusion matrix shows how the test examples were misclassified for each class. We observe that the majority of misclassifications happen for classes adjacent to the real class, that is, if the real difficulty of a text is C1, it is more likely that it is misclassified as B2 or C2 rather than as one of the other classes.


\begin{table}[h]\centering
\caption{Accuracy comparison across the various models for the task of foreign language difficulty prediction, and also baseline accuracy for a classifier predicting always the over-represented class.} \label{tab:accuracy}
\begin{tabular}{|l||c|c|c|}
 \hline
 & \texttt{LjL} & \texttt{sentencesBooks} & \texttt{sentencesInternet} \\ \hline
\textit{Baseline} & 0.32 & 0.17 & 0.17 \\ \hline
\textbf{CamemBERT \cite{martin-etal-2020-camembert}} & \textbf{0.79} & 0.86 & 0.53 \\
\textbf{GPT-3 Curie fine-tuned} & 0.76 & 0.89 & 0.61 \\
\textbf{GPT-3 DaVinci fine-tuned} & 0.77 & \textbf{0.90} & \textbf{0.62} \\ \hline
\end{tabular}

\end{table}




\section{Related work}

We aim to contribute to the literature on the use of computer-assisted tools in language learning or education in general. In this paper, we specifically focus on the difficulty estimation of foreign language texts, that is, on developing tools that can help language learners and teachers select appropriate reading material of specific topics and at specific difficulty level. Creating tools to assess the complexity of foreign language texts can assist learners and educators in choosing suitable materials that strike the right balance between challenge and comprehensibility, facilitating the gradual development of language proficiency.

One approach to estimating the difficulty of a foreign language text is to use readability formulas, which are mathematical algorithms that calculate the complexity of a text based on features such as sentence length and word frequency. Some commonly used readability formulas for foreign language texts include the Flesch-Kincaid Grade Level, the Simple Measure of Gobbledygook (SMOG), and the Gunning Fog Index. These technologies were initially developed for the English language, and gradually have been extended to other languages such as French, Chinese and Italian \cite{okinina2020ctap,chen2016ctap}. Note, that these readability formulas are primarily targeted to estimate the difficulty of a text for native speakers rather than for second language learners \cite{xia2019text}.

Another approach to difficulty estimation is to use machine learning techniques to analyze the text and predict its difficulty level based on various linguistic features. For example, researchers have used features such as syntactic complexity, word frequency, and semantic similarity to develop models that can accurately estimate the difficulty of foreign language texts \cite{chen2016ctap,franccois2012ai}.
A particularly notable advancement in this field in recent years is the integration of pre-trained word- and sentence- embeddings into text readability architectures \cite{wilkens2022fabra,imperial2021bert,lee2022editable,opencorporaFrench2022}. However, none of the previous works examines the predictive accuracy of difficulty estimation using the most recent GPT-3 large language models. As we show, such models can significantly boost the accuracy.

\section{Conclusion and next steps}

Several extant pedagogical approaches advocate the freedom of learners to follow their own, personalized and varied learning pathway, reading on topics they are passionate about. The reason for this approach is to make sure that reading in a foreign language becomes a daily habit and is not done as a chore.

We presented our solution of an educational content-based recommendation engine that helps learners discover appropriate content in a foreign language and presents content that is on par with their knowledge of the foreign language. Our solution crawls Internet content (text and video), and matches content with users based on their preferences and their knowledge of the foreign language. This solution requires algorithms that can a) predict the difficult level of text, b) predict the topic that the text represents and c) a recommender engine that matches content to users based on the users' interests and proficiency in a foreign language. In this paper, we specifically focused on one of the ingredients of the proposed application, the prediction of the difficult of the content using its text representation. To do so, we draw on the latest research in NLP, i.e., large language models and fine-tuned large language models for the goal of difficulty estimation. The fine-tuned models demonstrated that the difficulty estimation technique presented is significantly more accurate than existing and widely-used readability metrics. To the best of our knowledge, our work is the first to use GPT-3-based generative neural networks to create very accurate difficulty estimation models for foreign language text.

As next steps, we will enhance the content-based recommendation platform to use collaborative filtering methodologies \cite{koren2021advances}, i.e. using the collective historical preferences across like-minded users to discover appropriate personalised content. We also plan to conduct a large-scale evaluation of the tool at French-speaking universities in Europe to evaluate its effectiveness. Our aspiration is that with an extensive deployment, we can assess the effectiveness of machine learning solutions in accelerating the learning process of not only foreign languages but also of general cognitive skills.

\bibliography{bibliography}
\bibliographystyle{splncs04.bst}

\end{document}